# Transport Equation based Physics Informed Neural Network to predict the Yield Strength of Architected Materials

Akshansh Mishra
School of Industrial and Information
Engineering, Politecnico Di Milano,
Milan, Italy
akshansh.mishra@mail.polimi.it

*Abstract*— In this research, the application of the Physics-Informed Neural Network (PINN) model is explored to solve transport equation-based Partial Differential Equations (PDEs). The primary objective is to analyze the impact of different activation functions incorporated within the PINN model on its predictive performance, specifically assessing the Mean Squared Error (MSE) and Mean Absolute Error (MAE). The dataset used in the study consists of a varied set of input parameters related to strut diameter, unit cell size, and the corresponding yield stress values. Through this investigation the aim is to understand the effectiveness of the PINN model and the significance of choosing appropriate activation functions for solving complex PDEs in real-world applications. The outcomes suggest that the choice of activation function may have minimal influence on the model's predictive accuracy for this particular problem. The PINN model showcases exceptional generalization capabilities, indicating its capacity to avoid overfitting with the provided dataset. The research underscores the importance of striking a balance between performance and computational efficiency while selecting an activation function for specific real-world applications. These valuable findings contribute to advancing the understanding and potential adoption of PINN as an effective tool for solving challenging PDEs in diverse scientific and engineering domains.

## I. INTRODUCTION

Architected cellular materials are composed of interconnecting struts and nodes, and are also referred to as lattice structures or lattices. Another variety of these materials is minimum surfaces, which include continuous curved structures or sheets of curvature without clear nodes. Beyond the choice of base material (metallic, ceramic, polymeric, or composites) and the macroscopic part's shape (such as a femoral stem), the micro-architecture of these cellular materials gives designers an extra degree of customization [1-5]. Cellular materials function as structures on a microscopic level, while on a macroscopic level, they show traits of homogeneous materials. This indicates that the lattice and macroscopic scales are clearly separated by length scales. Since its behavior depends on the base material and the unit cell's design parameters, effective homogenous qualities, like an effective elastic modulus, can be used to describe it on a macroscopic scale. It is possible to span a variety of material properties, including stiffness, strength, torsion resistance, density, permeability, and thermal conductivity, among others, by varying the geometry of the unit cell while maintaining the base material [6-11]. The term "meta-materials" is now used to refer to all architected cellular materials because of this special capacity to customize the effective material properties.

A paradigm change in the creation of architected materials may be possible thanks to Physics-Informed Neural Networks (PINNs), which have special qualities. PINNs provide an effective and data-driven method to investigate the enormous material design space by utilizing the capabilities of neural networks and combining physics-based equations. As a result, architects' constructions can be quickly and precisely optimized without the need for time-consuming trial. With the help of complex interactions and behaviors revealed by PINNs, materials can be tailored to have the necessary qualities while still performing at their best for a variety of applications. The capability of PINNs to manage multi-physics interactions further increases their adaptability in many engineering contexts.

This is the first research work to implement the PINN model to evaluate the mechanical property of the BCC lattice based architected material. The primary goal of this research is to explore the utilization of the Physics-Informed Neural Network (PINN) model in tackling transport equation-based Partial Differential Equations (PDEs). The specific focus lies in evaluating the impact of various activation functions incorporated into the PINN model on its predictive capabilities, with a keen emphasis on assessing the Mean Squared Error (MSE) and Mean Absolute Error (MAE). To conduct this analysis, a diverse dataset comprising input parameters such as strut diameter, unit cell size, and corresponding yield stress values is utilized. Through this investigation, the study aims to gain valuable insights into the effectiveness of the PINN model and the significance of selecting appropriate activation functions for optimizing predictive performance in solving complex PDEs.

## II. MATHEMATICAL FORMULATION OF TRANSPORT EQUATION

This present research seeks to enhance the capabilities of the Physics Informed Neural Network (PINN) by integrating a Transport Equation based Partial Differential Equation (PDE) within its loss function. By doing so, the researchers aim to empower the neural network with a profound understanding of the underlying physical phenomena it models. The PDE's inclusion in the loss function facilitates the assimilation of domain-specific physics knowledge during the network's training, leading to superior accuracy and generalization in tackling intricate problems within the target domain. This innovative approach effectively bridges the gap between conventional physics-based modeling and contemporary deep learning techniques, presenting promising prospects for advanced predictive modeling and simulations across a diverse range of scientific and engineering disciplines.

The simplest form of the Transport Equation is shown in Equation 1.

$$u_t + u_x = 0 \ for \ u = u(x,t), x \in \mathbb{R}, t \in \mathbb{R} \qquad (1)$$

Where x is a one-dimensional spatial variable (can be n-dimensional also for e.g., $x \in \mathbb{R}^n$) and t is a time variable.

The more general form of the Transport Equation can be represented by Equation 2.

$$u_t + \bar{b}.D_x u = f \ for \ u = u(x_1, \ldots x_n, t), \bar{x} = (x_1, \ldots x_n) \in \mathbb{R}^n, t \in \mathbb{R} \qquad (2)$$

Where u which represents the mass density of a substance contained into a fluid and transported by fluid itself ($u = \rho(x,t)$) depends on (n+1) variables, $\mathbb{R}^n$ is the n-dimensional Euclidean space, $\bar{b}$ is a fixed vector and can be thought of as a velocity of fluid which is constant in time and f represents the sources which supply the substance into the region occupied by the fluid as shown in Figure 1.

The principal equation can be derived by imposing that the mass is preserved i.e., which obeys the law of conservation of mass for every Ω (sub-domain) contained into the region occupied the fluid we have as shown in Equation 3.

$$\frac{d}{dt}\int_\Omega \rho.dx = \int_\Omega f.dx - \int_{\partial\Omega} \rho.\bar{v}.\gamma \qquad (3)$$

Equation 3 can be subject to additional manipulations and solutions, as demonstrated in Equations 4 and 5 which represents the Divergence theorem or Gauss-Green theorem. Furthermore, it can be reformulated within the context of the conservation of mass equation as shown in Equation 6.

$$\frac{d}{dt}\int_\Omega (\rho(x,t).dx) = \int_\Omega \rho_t(x,t).dx \qquad (4)$$

$$\int_{\partial\Omega} \rho.\bar{v}.\gamma = \int_{d\Omega} X.\gamma = \int_\Omega div.X \qquad (5)$$

$$\int_\Omega \rho_t = \int_\Omega f - \int_\Omega D_x.\rho.b \Rightarrow \int_\Omega \rho_t + \int_\Omega D_x.\rho.b = \int_\Omega f \qquad (6)$$

The Cauchy problem, an initial value problem, is utilized to explore the evolution of mass density represented by the function "u" from its initial data as shown in Equation 7.

$$\begin{cases} u_t + b.D_x u = f, x \in \mathbb{R}^n, t > 0 \\ u(x,0) = g(x), x \in \mathbb{R}^n \end{cases} \qquad (7)$$

Figure 2 presents both the surface plot and contour plot of the Transport equation, providing a comprehensive visual representation of the equation's behavior. These graphical illustrations offer valuable insights into the complex dynamics and spatial variations of the Transport equation, enhancing our understanding of its underlying properties.

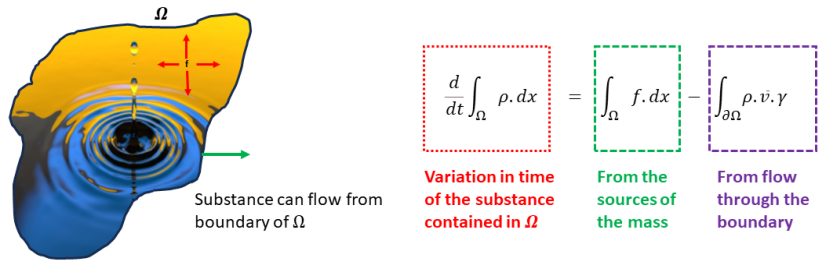

*Figure 1. Schematic representation of the mathematical formulation of Transport equation based PDE*

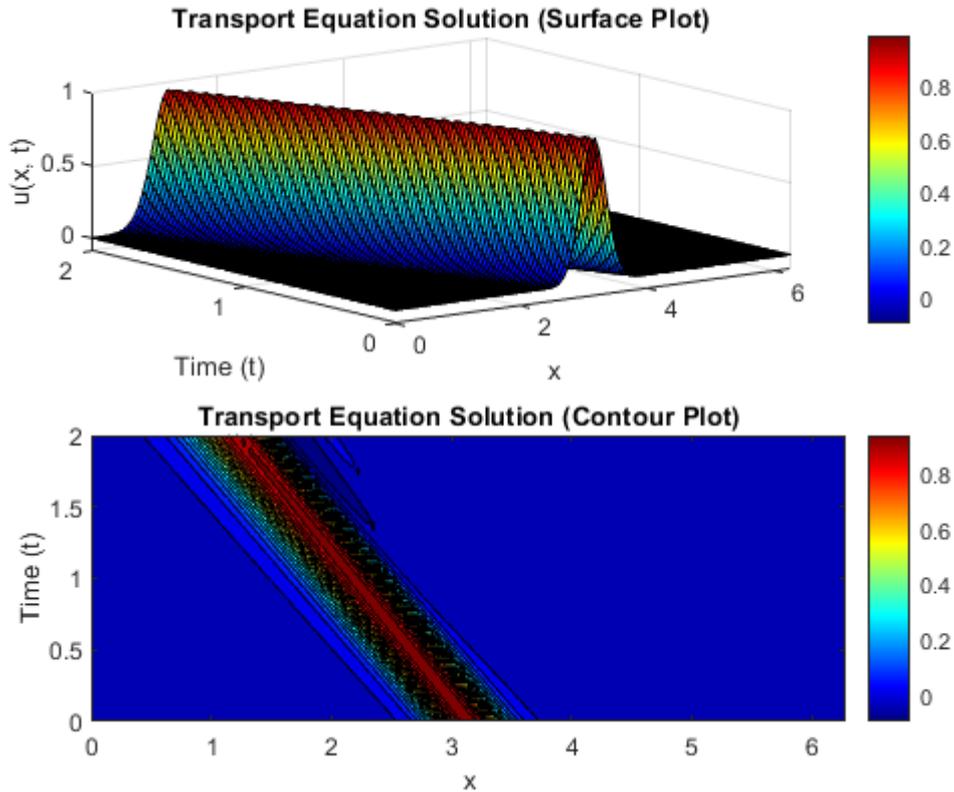

*Figure 2. Surface and Contour plots based on the solution of Trnasport Equat*

## III. MECHANISM OF ACTIVATION FUNCTIONS IN NEURAL NETWORKS

In modern neural networks, activation functions play a critical role in enabling the learning of complex patterns from data. Most activation functions are non-linear as shown in Figure 3, which allows neural networks to capture intricate features present in datasets, such as recognizing different pixels that make up an image's specific feature. By introducing non-linearity, these activation functions make it possible for neural networks to learn from a wide range of input data, enabling them to perform tasks that require understanding complex and non-linear relationships. In essence, activation functions determine whether a neuron should "fire" or be activated based on the weighted sum of its inputs. Figure 4 shows the overall comparison of the activation functions.

The significance of non-linear activation functions lies in their ability to go beyond merely learning linear and affine functions. Without non-linear activation functions, neural networks would be limited to approximating linear combinations of input features, which is insufficient for capturing the intricate relationships and patterns that are common in real-world data. Linear and affine functions lack the expressive power needed to represent the complexities present in various datasets, such as those encountered in natural images, speech, or natural language. To illustrate this issue, consider a simple example of binary classification. If we attempt to classify points on a 2D plane into two classes using only a linear model, the decision boundary would be a straight line. However, many real-world datasets require decision boundaries that are more complex, possibly curved or irregular, to separate different classes accurately. Non-linear activation functions, like ReLU, sigmoid, or tanh, introduce the required non-linearity in the model, allowing it to learn intricate decision boundaries and patterns that linear functions cannot capture as shown in Figure 5.

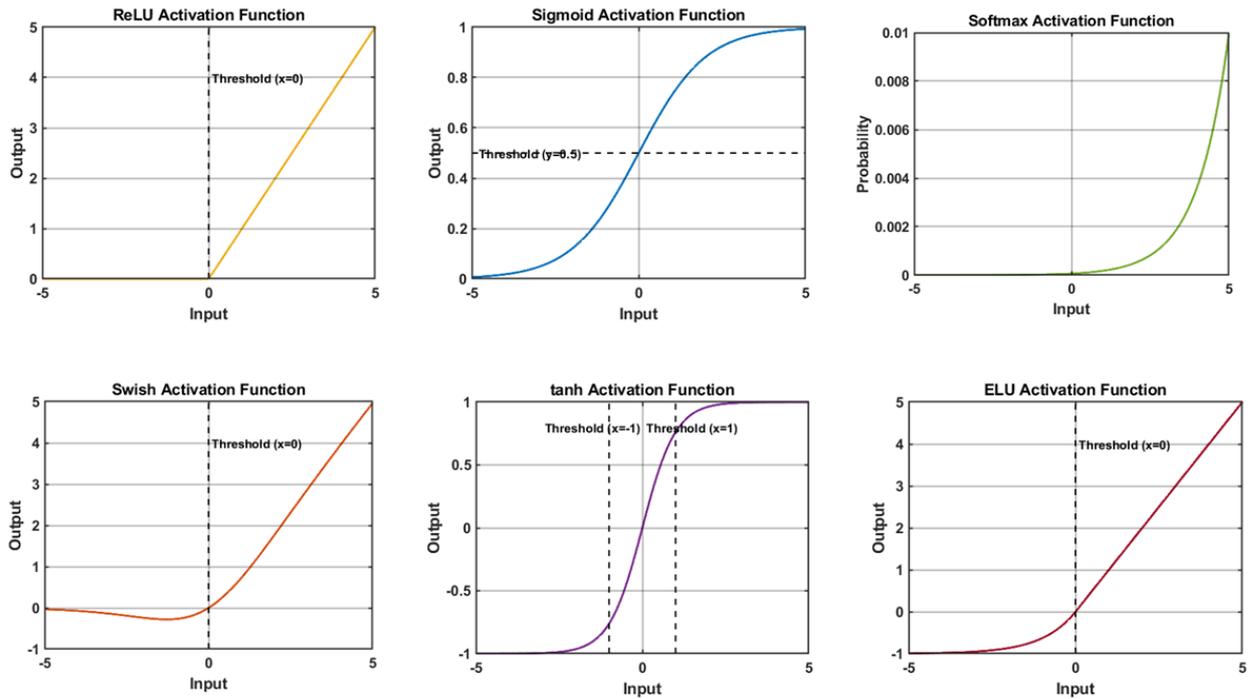

*Figure 3: Plots of various activation functions. The ReLU activation function is one of the most widely used activation functions in deep learning. It takes the input value and returns the input itself if it is positive, or zero if the input is negative (i.e., max(0, x)). This introduces non-linearity and helps the network learn complex representations. However, ReLU can cause the "dying ReLU" problem where some neurons become inactive during training, leading to the loss of gradient flow and hampering learning. The tanh activation function maps the input to the range (-1, 1). It is zero-centered, which mitigates the vanishing gradient problem associated with the sigmoid function. However, tanh is still susceptible to the vanishing gradient problem for extreme inputs. The sigmoid activation function maps the input to the range (0, 1). It is commonly used in the past for binary classification problems. However, sigmoid has vanishing gradient problems, making it less suitable for deep networks. The ELU activation function is another alternative to ReLU that addresses the "dying ReLU" problem. For negative inputs, it introduces an exponential component to produce negative values and allows non-zero gradients. The Swish activation function combines the linearity of ReLU for positive inputs with the smoothness of the sigmoid function for negative inputs. This hybrid function is differentiable and has shown promising performance in some scenarios.*

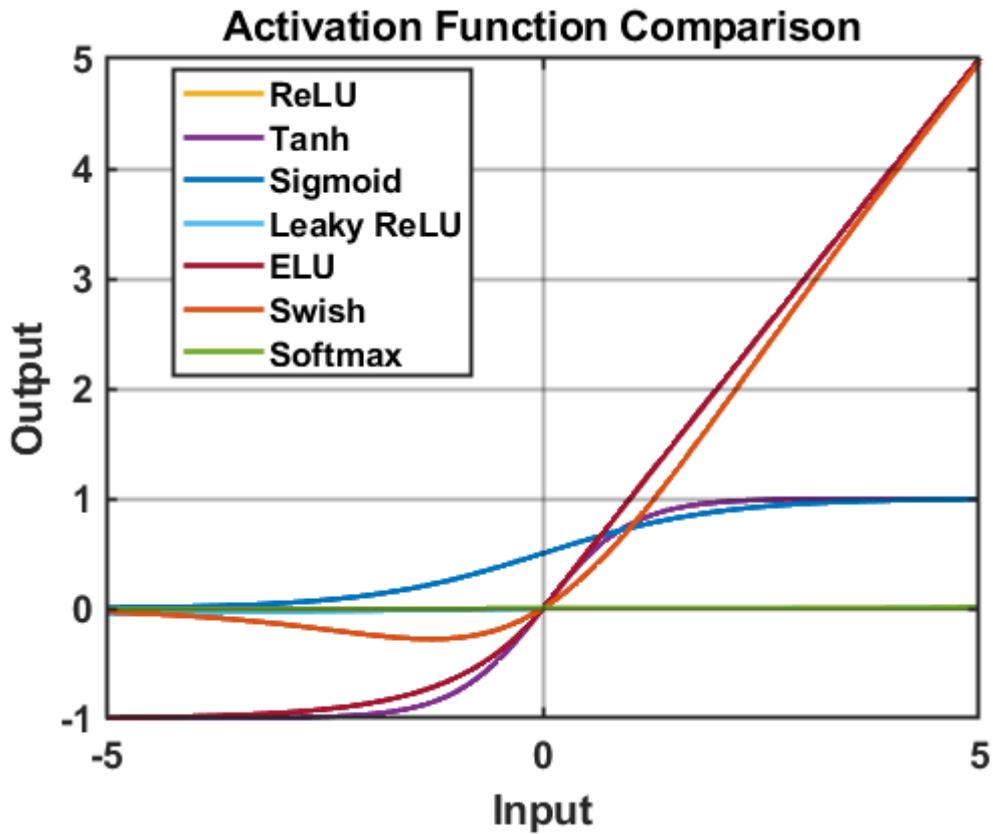

*Figure 4. Comparison of the plots of the activation functions.*

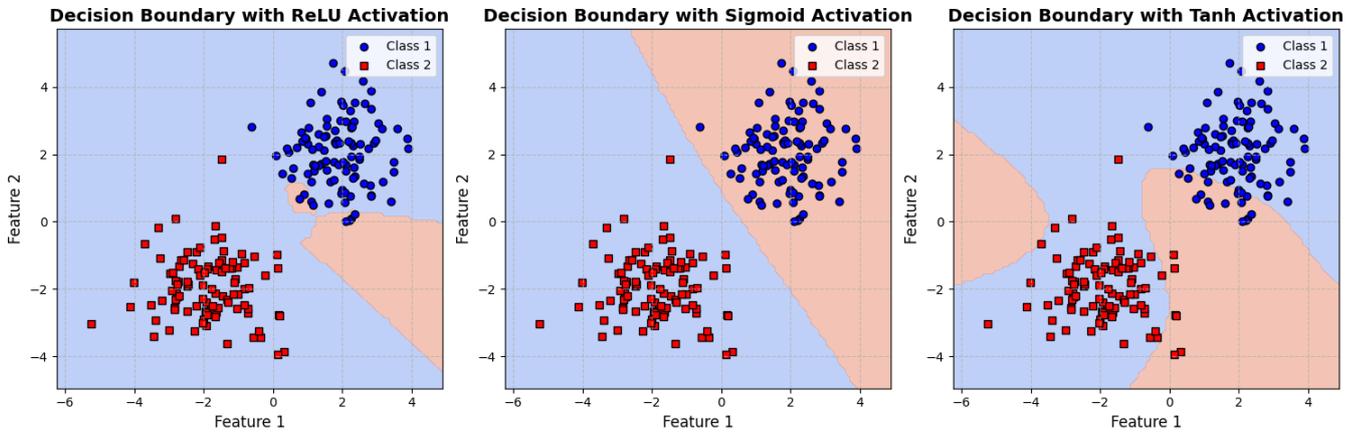

*Figure 5. Comparison of Decision Boundaries with Different Activation Functions for Binary Classification. his figure showcases a comprehensive comparison of decision boundaries obtained using various activation functions in the context of binary classification. The study employs a Multi-Layer Perceptron (MLP) classifier with one hidden layer containing 10 neurons to investigate the learning capacity of neural networks under different activation functions. The synthetic dataset comprises two distinct classes, denoted by blue circles (Class 1) and red squares (Class 2), each consisting of 100 data points. These points are strategically distributed to challenge the classifiers in learning non-linear decision boundaries effectively.*

IV. EXPERIMENTAL PROCEDURE

In this study, the experimental data was obtained from the research conducted by Azmi et al. [12]. They used Fused Deposition Modeling (FDM) Additive Manufacturing (AM) technology on a CubePro machine to create lattice structure samples from acrylonitrile butadiene styrene (ABS) thermoplastic material. The lattice samples had a cubic shape with dimensions of 60 mm³ and were designed using a combination of parameters in SolidWorks computer-aided design (CAD) software. These designs were then saved as .STL files and imported into the CubePro software for the slicing process.

Figure 6 illustrates the 60 mm³ tessellated Body-Centered Cubic (BCC) lattice structure and a 20 mm BCC unit cell with a 5 mm strut diameter.

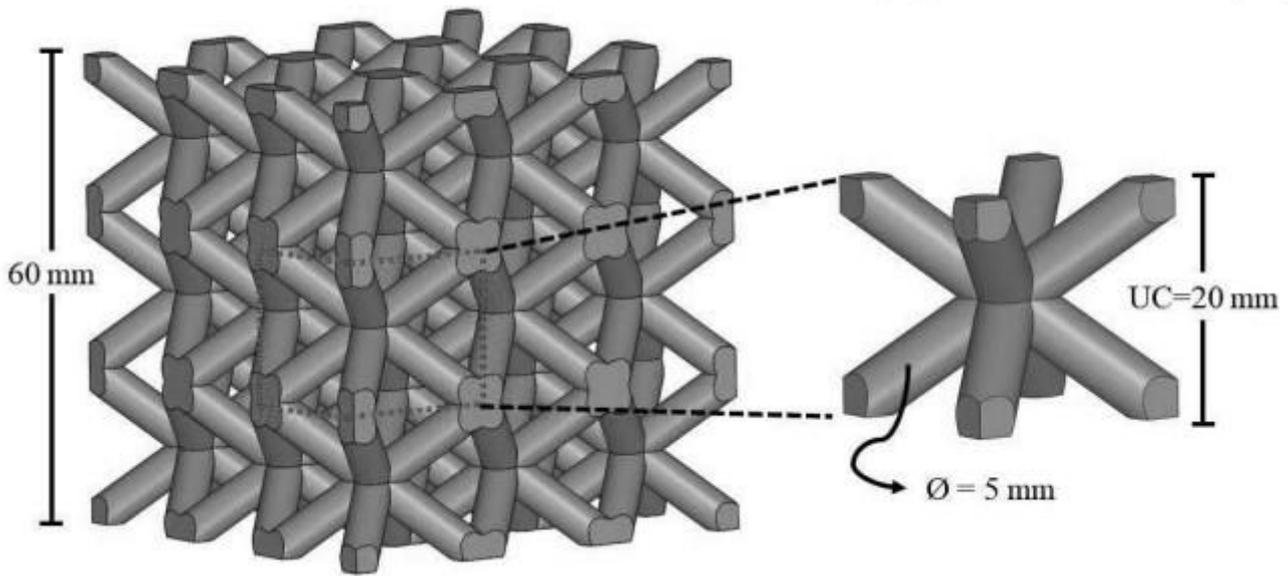

*Figure 6. A lattice structure featuring a unit cell size of 20 mm and struts with a diameter of 5 mm [12]. The application area of lattice structures with Body-Centered Cubic (BCC) lattice strut-based designs is diverse. These structures find applications in various fields, including aerospace engineering, where they are used for lightweight structural components, thermal management solutions, and load-bearing structures. They are also employed in medical implants, as their porous nature allows for enhanced osseointegration and nutrient diffusion.*

The mechanism used in this research work for synthetic data generation is polynomial regression and interpolation. Polynomial regression is a type of linear regression where the input features are transformed by raising them to certain powers. In this research, the PolynomialFeatures class from scikit-learn is employed to transform the input parameters into polynomial features of a specified degree (in this case, degree=3). This allows the model to capture more complex relationships between the inputs and the output. After fitting the polynomial regression model, the code proceeds with interpolation. It generates a new dataset by interpolating the existing data points to create additional data points that lie between the original data points. This is achieved using the np.interp function in NumPy, which performs linear interpolation. The original dataset contains a limited number of data points, but interpolation allows us to estimate the output parameter for a much larger number of input parameter values, thereby creating synthetic data as shown in Figure 7.

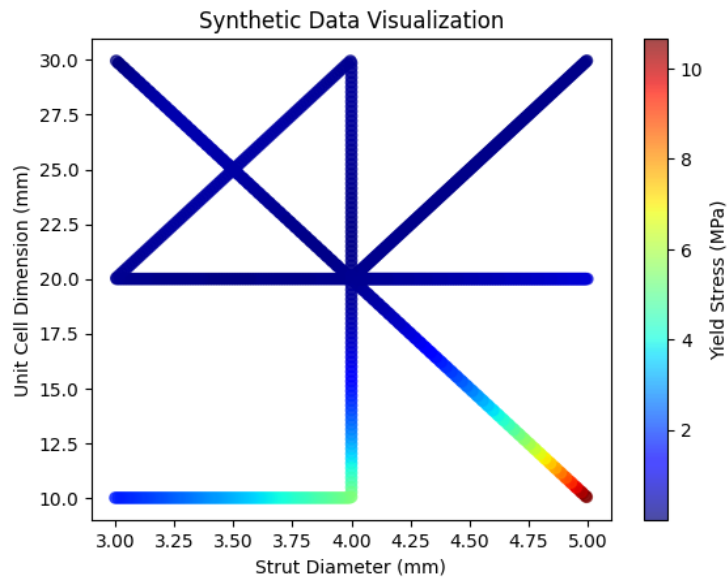

*Figure 7. Visualization of the generated synthetic 1000 data points in the present work.*

In the present work, strut diameter in mm and unit cell dimensions in mm are considered as input parameters while the yield stress in MPa is considered as an output parameter. The PINN model architecture is implemented using TensorFlow and Keras libraries. The neural network consists of multiple dense layers with tunable hidden units. Various activation function such as elu, relu, swish, sigmoid etc. have been used in the present work and their influence on the predicted output that is the Yield Stress of the BCC lattice structure will be seen in results and discussion section.

The PINN model incorporates the transport equation into the loss function as shown in Figure 8 which is the framework used in the present work. The transport equation describes the advection process in the system and involves spatial and temporal derivatives. By computing the derivatives of the model predictions with respect to spatial and temporal inputs, the PINN model enforces the underlying physics within the learning process. The dataset is split into training and testing sets to validate the PINN model's performance. The model is trained using an optimization algorithm, specifically the Adam optimizer, for a specified number of epochs. During each epoch, the PINN model minimizes the combined loss, which includes both the physics-informed loss and the data-driven loss (mean squared error between predictions and actual outputs). After training the PINN model, its performance is evaluated using evaluation metrics such as Mean Squared Error (MSE), and Mean Absolute Error (MAE) values. These metrics quantify the accuracy and predictive capability of the PINN model in approximating the transport equation-based PDE solutions.

During training and testing, we will record and compare the MSE and MAE for each activation function used in the PINN model. The model's convergence speed, accuracy, and stability will be evaluated based on these metrics. By comparing the performance of the PINN model with different activation functions, we can identify the most suitable activation function for the specific transport equation-based PDE problem.

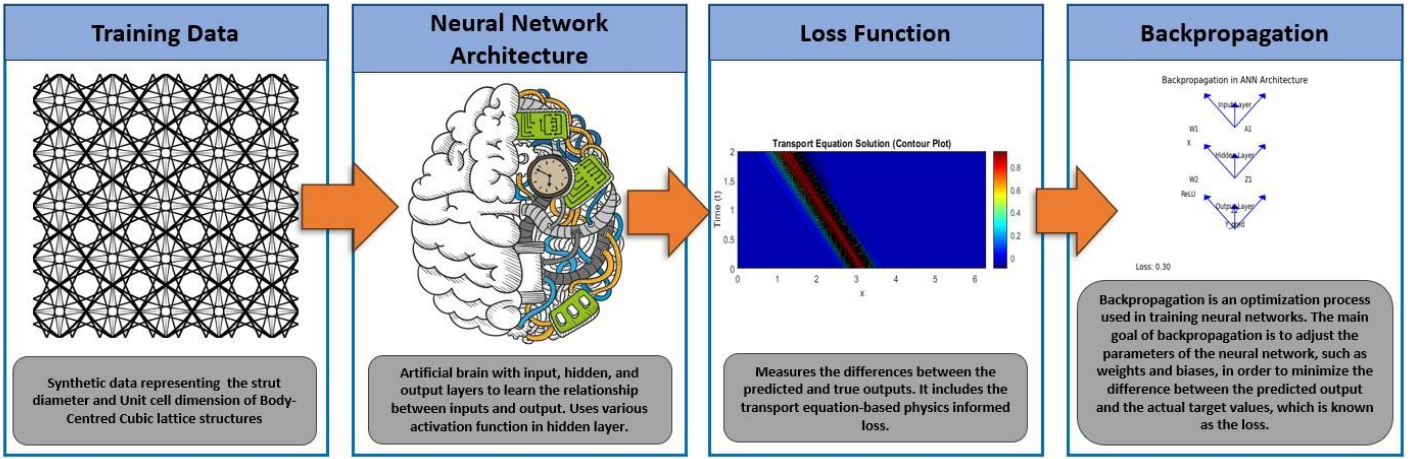

*Figure 8. PINN framework used in the present work. The PINN model incorporates the transport equation into the loss function. The transport equation describes the advection process in the system and involves spatial and temporal derivatives. By computing the derivatives of the model predictions with respect to spatial and temporal inputs, the PINN model enforces the underlying physics within the learning process.*

## V. RESULTS AND DISCUSSION

The fundamental essence of the PINN model lies in integrating the transport equation directly into the loss function. In this context, the transport equation is assumed to be represented by Equation 8. By doing so, the PINN model gains the ability to learn and optimize its parameters based on the governing physics encoded in the equation, which contributes significantly to its accurate predictive capabilities.

$$\frac{du}{dt} + c \cdot \frac{du}{dx} = 0 \qquad (8)$$

where "u" is the "Yield Stress," "t" represents time, "x" represents spatial coordinates, and "c" is the advection velocity. The PINN model approximates "u" as a function of "x" and learns the value of "c" that minimizes the error between the left-hand side and the right-hand side of the equation.

The loss function in the PINN model comprises two components: the physics-informed loss and the data-driven loss. The physics-informed loss enforces the transport equation, while the data-driven loss measures the difference between the predicted and actual "Yield Stress" values.

Table 1 shows the obtained metric features with respect to each activation function trial individually. The role of the activation function in the PINN model is critical as it enables the model to capture non-linear relationships and solve complex PDEs accurately. By comparing different activation functions in terms of their impact on MSE and MAE, we can make an informed choice to enhance the PINN model's performance for the specific problem at hand. The selection of an appropriate activation function contributes significantly to the PINN model's success in accurately predicting physical behaviors and solving challenging PDEs.

Table 1 show the obtained metric features value obtained for each activation functions used in the present work for a comparative study. Figure 8 shows the respective MSE and MAE plots for each activation functions.

Table 1. Comparison of metric features with respect to the activation functions

| Activation Function | MSE | MAE |
|---|---|---|
| Relu | 4.393951 | 1.658419 |
| Tanh | 4.396555 | 1.658940 |
| Sigmoid | 4.370328 | 1.653357 |
| Leaky Relu | 4.378629 | 1.655049 |
| Exponential Linear Unit | 4.366074 | 1.652316 |
| Swish | 4.395654 | 1.658550 |

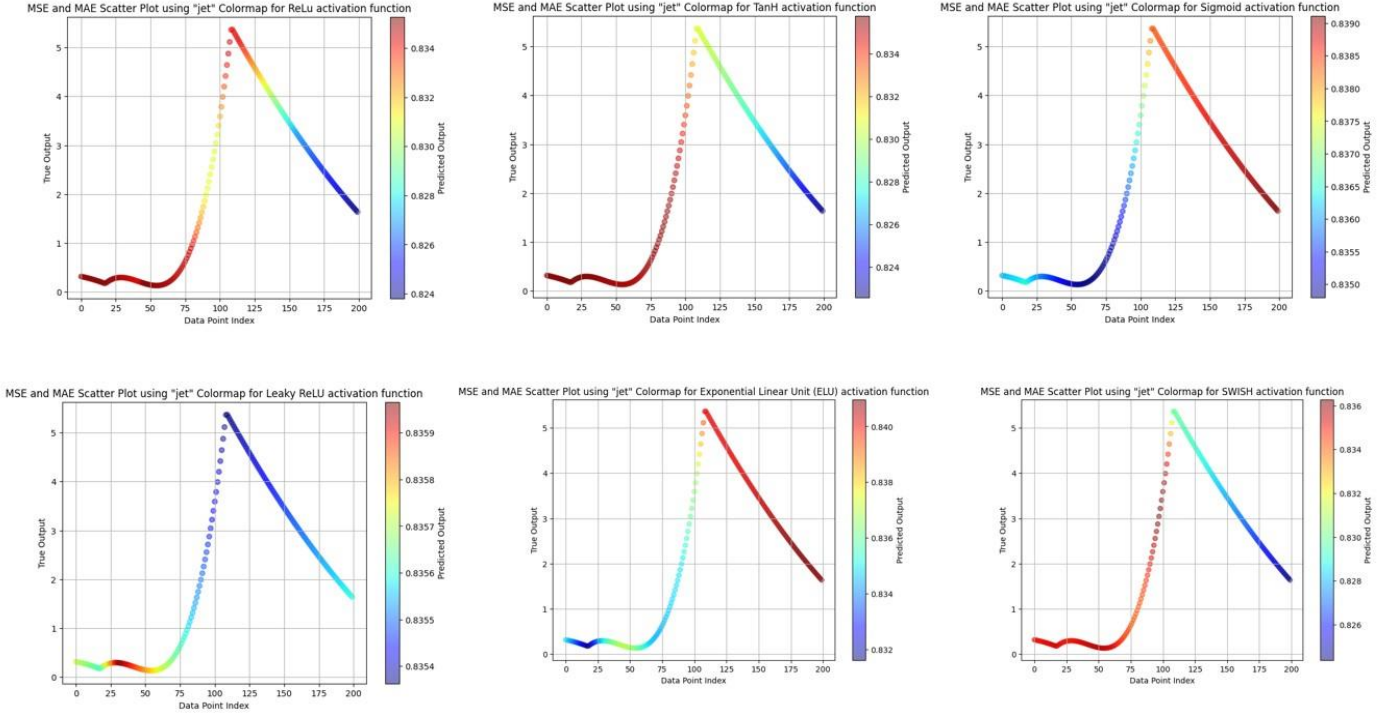

*Figure 9. MSE and MAE plots for showing the variation of metric features corresponding to the given activation functions.*

The findings from the results indicate that all the tested activation functions are appropriate for addressing this specific problem. Although some activation functions might demonstrate better performance under certain conditions, the slight variations in MSE and MAE suggest that any of these activation functions can effectively be employed. However, it is crucial to consider computational efficiency as another significant factor apart from performance. Despite the minimal disparities in MSE and MAE, some activation functions might impose lower computational demands, making them more appealing for large-scale or resource-intensive applications. Therefore, researchers and practitioners should take into account both performance and computational efficiency when selecting an appropriate activation function for their particular use case.

## VI. CONCLUSION

The research undertaken provides an extensive investigation into the effectiveness of the Physics-Informed Neural Network (PINN) model for addressing transport equation-based Partial Differential Equations (PDEs). The main focus was on evaluating the impact of different activation functions within the PINN model and their influence on performance measures, such as Mean Squared Error (MSE) and Mean Absolute Error (MAE). The dataset consisted of diverse input parameters concerning strut diameter, unit cell size, and corresponding yield stress values. Through a rigorous experimental analysis, the study examined six widely used activation functions, including ReLU, Tanh, Sigmoid, Leaky ReLU, Exponential Linear Unit, and Swish. Surprisingly, the results indicated that all activation functions demonstrated remarkably similar MSE and MAE values, signifying consistent and precise predictive capabilities across the entire spectrum of activation functions. This remarkable consistency in performance underscores the PINN model's exceptional ability to effectively capture the underlying physics and comprehend intricate data relationships. The study acknowledged that there are additional factors to be considered beyond performance metrics. Of particular importance is computational efficiency, especially in scenarios involving large-scale or resource-intensive applications. Although the differences in MSE and MAE were minimal, some activation functions exhibited lower computational overhead, making them more suitable for situations where efficiency is a critical concern.